\documentclass{article} 
\usepackage{url}
\usepackage{xcolor}     
\usepackage{tcolorbox}  
\usepackage{graphicx}
\usepackage{caption}    
\usepackage{amsmath}
\usepackage{amssymb}
\usepackage{algorithm}
\usepackage{algpseudocode}
\title{The Ann Arbor Architecture\\for Agent-Oriented Programming}
\author{Wei Dong \\ Ann Arbor Algorithms \\ \texttt{wdong@aaalgo.com}}
\date{February 14, 2025}

\begin{document}
\maketitle
\begin{abstract}
In this paper, we reexamine prompt engineering for large language models through the lens of automata theory. We argue that language models function as automata and, like all automata, should be programmed in the languages they accept, a unified collection of all natural and formal languages. Therefore, traditional software engineering practices—conditioned on the clear separation of programming languages and natural languages—must be rethought. We introduce the Ann Arbor Architecture, a conceptual framework for agent-oriented programming of language models, as a higher-level abstraction over raw token generation, and provide a new perspective on in-context learning. Based on this framework, we present the design of our agent platform Postline, and report on our initial experiments in agent training.    
\end{abstract}

\section{Introduction}

\subsection{Large Language Models Are Automata}

The foundation of modern computing was established by early pioneers striving to formalize computation through various theoretical models, with automata theory becoming a central framework \cite{hopcroft1979}. Alan Turing’s introduction of the Turing machine\cite{turing1936} provided a universal model for computation, employing an imaginary tape as a conceptual memory device and programming medium. This focus on theoretical machines and symbol manipulation ultimately culminated in the unification of automata and formal languages, exemplified by Noam Chomsky’s classification of formal languages\cite{chomsky1956}, and firmly established the role of languages in computer science.

Building on this foundation, software engineering evolved as a discipline dedicated to creating higher-level abstractions that enhance programming efficiency and organization. Programming languages progressed from machine code to assembly, structured programming and beyond, each serving as a virtual machine that facilitates the translation of code into executable instructions for a Turing-equivalent system.  The adoption of design patterns and software engineering methodologies provided structured approaches to managing complexity and improving software development workflows. While these methodologies did not extend the fundamental computational capabilities of the Turing machine, they optimized the way computational processes are expressed, organized, and refined.

\begin{figure}[h]
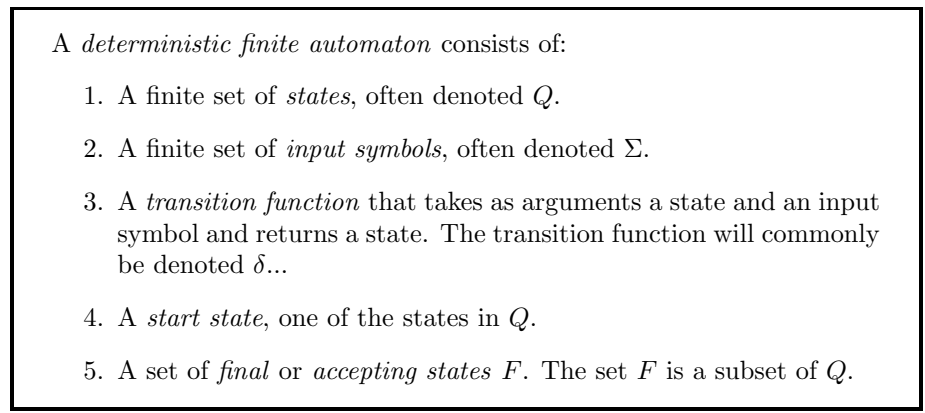

    \centering
    \begin{tcolorbox}[width=1.0\linewidth, colframe=black, colback=white, sharp corners]
        A \emph{deterministic finite automaton} consists of:
        \begin{enumerate}
            \item A finite set of \emph{states}, often denoted $Q$.
            \item A finite set of \emph{input symbols}, often denoted $\Sigma$.
            \item A \emph{transition function} that takes as arguments a state and an input symbol and returns a state. The transition function will commonly be denoted $\delta$...
            \item A \emph{start state}, one of the states in $Q$.
            \item A set of \emph{final} or \emph{accepting states} $F$. The set $F$ is a subset of $Q$.
        \end{enumerate}
    \end{tcolorbox}
    \caption{Definition of DFA by Hopcroft, Motwani and Ullman\cite{hopcroft2006}.  With language models, a \emph{symbol} is commonly called a \emph{token} and the \emph{alphabet} the \emph{vocabulary}.}
    \label{fig:quote}
\end{figure}

Large language models, such as ChatGPT, represent a major advancement in artificial intelligence, capable of processing and generating text like humans do across a broad array of topics. Although they operate atop an extensive computational infrastructure and draw upon vast resources, basic language models -- excluding reasoning-augmented models like GPT-o1 -- still largely conform to the original definition of an automaton (Figure~\ref{fig:quote}): they read input tokens, update internal hidden states, and produce output tokens.

A defining feature of large language models is their ability to process both natural and formal languages. Unlike conventional compilers and interpreters which are limited to predefined programming languages, large language models seamlessly handle diverse linguistic structures, spanning human languages, mathematical expressions, and code. Just like an automaton is programmed in the language it accepts, we argue that {\bf a large language model should be programmed in the same languages it processes}, that is, the full range of natural and formal languages. This paradigm shift necessitates rethinking software engineering methodologies to maximize the potential of large language models as a computing platform.

\subsection{Agent-Oriented Programming}

From the perspective of software engineering as discussed above, language models operate at the abstraction level of finite automata, or programmable virtual machines. While they can produce coherent text and engage in sophisticated interactions, their responses remain bound to token-level transitions. Just as programming methodologies have enabled the growth of software engineering, higher-level abstractions will unlock greater potential in language models. The full scope of language model programming is still unfolding, being developed in ways that are largely unconcious, and overshadowed by the dominant efforts aimed at improve model reasoning through machine learning approaches\cite{openai24b,deepseek25}. Among the various emerging directions, many point to agents as a promising paradigm for organizing and managing language model-based applications\cite{masterman24,wang24,google24}. We too consider agents of foundational importance and advocate for the term \textbf{agent-oriented programming} as a formal methodology, following the tradition of structured, object-oriented, and functional programming.

An agent is a natural progression beyond language models, intended to endow AI systems with more human-like behavior. While existing survey papers often assume the value of agents with minimal explanation, we take an epistemological perspective to underscore why agents are critical in successful application of language models. Language models possess vast amounts of knowledge, but due to the diverse and sometimes contradictory data on which they are trained, these models do not hold a consistent worldview, set of values, or unified objectives, rendering them sometimes too flexible and unreliable for tasks requiring stable perspectives. In essence, an language model functions like an ``average" human who may exhibit contradictory views, rather than a distinct individual with a coherent perspective.

The concept of an agent addresses this limitation by filtering the language model’s vast knowledge to emphasize only the subset relevant to a particular individual. The agent remains aware of the broader knowledge encoded within the model but, in cases of conflicting perspectives, maintains a defined stance, ensuring consistency and reliability.
Because an agent narrows the capacity of an raw language model, it naturally follows that an agent-based  platform must support multiple agents to compensate for the loss of generality in any single agent. The platform must facilitate efficient information exchange between agents, effective conflict resolution, and dynamic adaptation to evolving objectives. The different agents adopt distinct perspectives or roles and collectively leverage the full breadth of the language model's capabilities.

Many agent platforms have been developed.  Google's Vertex AI Agent Builder/Dialogflow~\cite{vertex24} and Microsoft's Copilot Studio\cite{azure24} are two leading enterprise platforms.  There are also popular open-source projects like AutoGen\cite{autogen} and CrewAI\cite{crewai}.  These platforms provide a variety of interfaces and approaches for building and managing agents, but they mostly constrain themselves to a common paradigm:
\begin{itemize}
    \item The user (designer or engineer) has to assume a specific task to be automated.  It is usually also assumed that the same task has to be done repeatedly.
    \item The user has to decompose the task into sub-tasks, each being easy enough to be solved by invoking language models.  These sub-tasks form a computation graph, either explicitly or implicitly.
    \item The user has to specify the prompts for each node of the computation graph.
    \item The end product of agent programming is in the form of a program or an online service that is separated from the design interface.
\end{itemize}
While such an approach does allow solving a plethora of real-life tasks, it inherently limits the flexibility and adaptability of language models.
Given the generality of language models, as previously discussed, we argue that static computation graphs -- or structured approaches in general -- are remnants of traditional software engineering practices from the age of formal computer languages. Relying on such rigid structures risks constraining the full potential of language models and preventing them from achieving more autonomous and contextually aware behaviors.  Specifically, we argue that the existing platforms do not allow the form of in-context learning as we discuss below.

\subsection{Rethinking In-Context Learning}

The rapid advancement of language models has been largely driven by scaling laws~\cite{kaplan20}, where continually increasing model size and training data has yielded steady gains in performance. However, recent evidence~\cite{reuters24} suggests that this strategy may be approaching its limits, as further scaling appears to offer diminishing returns.  
While vast efforts are currently dedicated to refining reasoning algorithms, we argue that the form of training data must also be reexamined. 

One major gap in current training paradigms is the absence of learning through multi-round interactions.
The standard training process, as exemplified by Llama 3\cite{llama24}, is to pre-train a model with a large amount of free text, and then to post-train the model to follow instructions using supervised training.  The post training examples follow a rigid structure, typically in the form of either (input, expected output) pairs for SFT \cite{wei01,sanh21,wang22} or (input, preferred output, rejected output) triplets for DPO (direct preference optimization)\cite{dpo24}. In the case of DPO, the rejected mistakes are not made by the AI being trained but are staged.  While these approaches provide clear guidance on the end goal, they lack the iterative, trial-and-error learning process fundamental to human skill acquisition. Effective training should involve opportunities for an AI model to make mistakes, receive feedback with explanations of why, and adjust its behavior accordingly -- mirroring how humans refine their skills under the guidance of a mentor.

Based on this observation, we propose to redefine \emph{in-context learning}  or \emph{in-context training} as an evolving process occurring through message exchanges that shape the agent's behavior. This approach extends beyond today's common practice of merely inserting a few shot of training samples into the prompt. Instead, the prompt should incorporate memory episodes of dynamic teacher-student interactions. A typical episode involves the presentation of a task, one or more erroneous attempts, the teacher’s correction and explanation to each of those, and ultimately, the correct solution with confirmation.  In our proposed framework, in-context learning occurs naturally, without a rigid division between training and deployment. As the conversation progresses, the process smoothly transitions from training to deployment.  During the agent's subsequent operation, additional in-context samples are picked up whenever outliers trigger human intervention.  

\subsection{Paper Organization}

Section 2 elaborates on the proposed Ann Arbor Architecture. Section 3 presents the design of Postline, our prototype platform. Section 4 presents a series of experiments on an early version of Postline. Section 5 provides further thinkings and concludes the paper.

\section{The Ann Arbor Architecture}
\subsection{Overview}

The Ann Arbor Architecture aims to create a framework for understanding agents, their connection with language models, and their engagement with tasks and environments.
It is a mind framework rather than the architecture of a realized system.
We challenge the current task-centric agent architectures by highlighting agents' collaboration with engineers, their persistence, long-term growth and even evolution.  
Our goal is to unify various engineering processes, a goal naturally called for by the language model's unification of natural and formal languages.

At its core, the proposed framework models agents and their interactions after the email system. Each agent is assigned a unique identifier—an email address—through which it exchanges messages. All messages with the agent's address  appearing in either the ``From" or the ``To"-like fields form a persistent, chronological record of the agent’s communication history, collectively referred to as the agent’s {\bf journal}. This persistent journal serves as the primary memory mechanism for the agent, ensuring that all past interactions remain accessible and can inform future decision-making.

Under the proposed framework, there is only one form of information exchange: sending emails.  This entails two important distinctions between the proposed framework and the existing approaches.

\begin{figure}[h]
    \centering
    \includegraphics[trim=0 0 0 0,clip,width=0.8\linewidth]{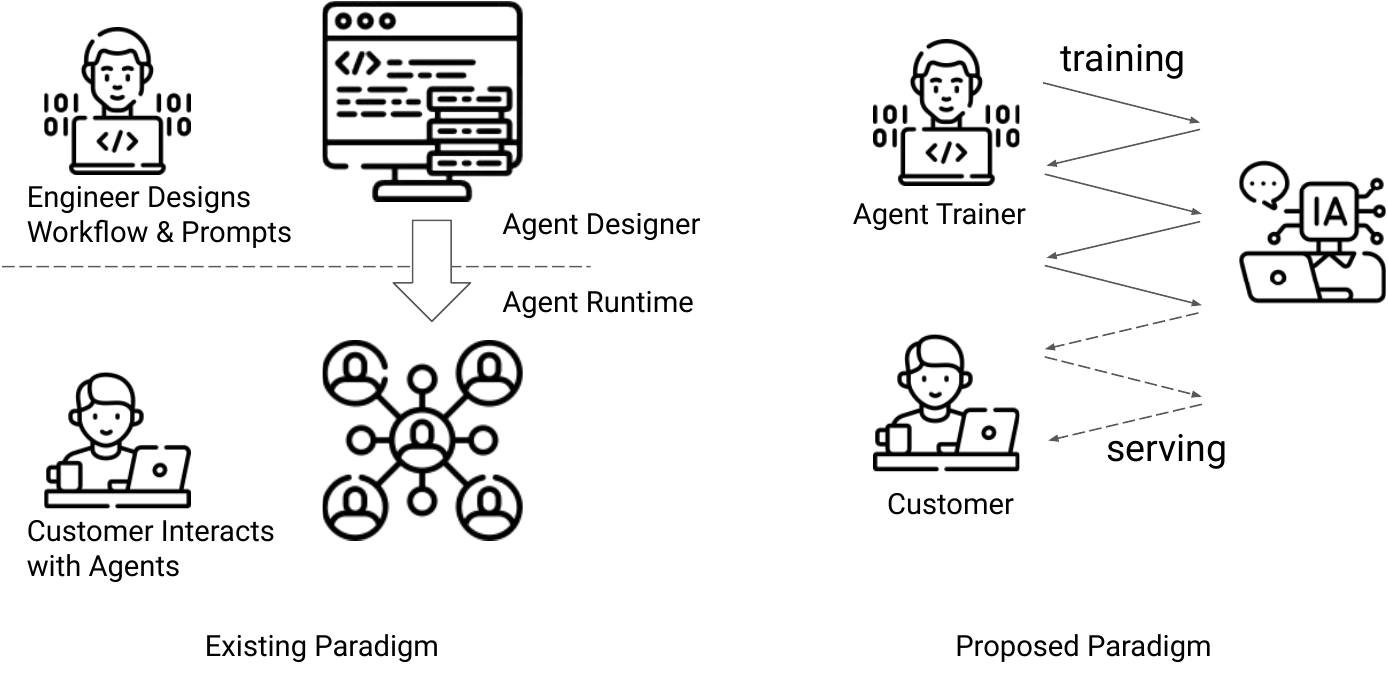}
    \caption{Different paradigms of customer service agents.  With existing platforms, development and deployment are separate processes with very different software environments.  Under the proposed framework, both training and serving occur in conversations on the same platform and the transition is seamless.}
    \label{fig:postline_flow}
\end{figure}

First, our proposed framework does not separate the design/training phase from the deployment phase. We illustrate this using the development of a customer service conversational agent(Figure~\ref{fig:postline_flow}). In existing systems, engineering and deployment occur in distinct stages: during engineering, developers design workflows and write prompts for different scenarios (the different nodes of a computation graph). Subsequently, in testing and deployment, the computation graph is installed on a (logical) state machine that uses these prompts with language models to drive state changes.

Under our proposed framework, however, this becomes a single continuous process. The trainer teaches the agent to handle customers by conducting email conversations. At some point, the trainer poses as a customer to test the agent’s capabilities. If the agent passes the tests, it naturally transitions to handling real customer interactions in emails (as an underlying communication protocol rather than the actual software interface the trainers and customers see).

\begin{figure}[h]
    \centering
    \includegraphics[trim=0 0 0 0,clip,width=0.8\linewidth]{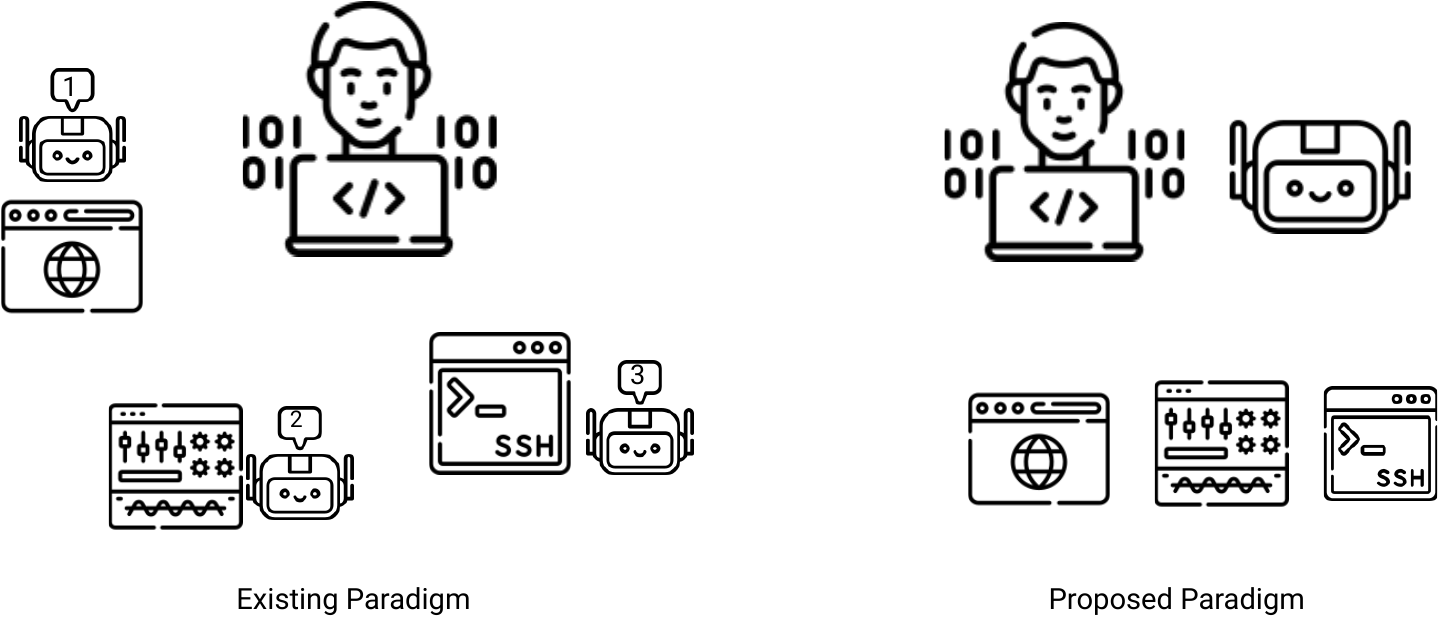}
    \caption{
    Different paradigms of research assistant agents.
    In the existing task-oriented paradigm, each tool has its own agent, and the researcher interacts with multiple agents that are unaware of each other. Under the proposed framework, a single agent operates across all tools to assist the researcher, maintaining continuous memory even when focus shifts between tools.  Each tool implements the non-intelligent \emph{robot} protocol so they can be automated by the agent.
    }
    \label{fig:postline_research}
\end{figure}

Second, our approach is to develop agents that are not confined to a single task or environment. We illustrate this with a research scenario involving assistant agents (Figure~\ref{fig:postline_research}). In existing systems, if a data scientist conducts research using a browser, a coding IDE like Cursor~\cite{cursor24}, and a terminal, a separate agent might be embedded in each of these tools to assist the user. These agents are typically developed by or for the provider of the respective tool.

Under our proposed framework, a single agent (or a team of agents) work across these tools to assist the researcher.  When the researcher switches the browser to the IDE, the same agent is present in the IDE with the recent memory of a browsing session. We propose repurposing the term {\bf robot} to refer to non-intelligent adapters that can be automated by an agent, as well as software and hardware entities equipped with such adapters.  So each tool appears as a robot with its own email address under our framework.

\subsection{Emails and the MBox Format}

The MBox format was first implemented in Fifth Edition of Unix and was standardized in RFC4155~\cite{rfc4155}.

Under the proposed framework, agent communication is modeled after email systems, utilizing the MBox format as the conceptual data format for both agent memory and agent communication.  Below is a sample MBox file containing two messages, separated by a blank line followed by a line starting with \texttt{From}.  A separating line starting with \texttt{From} is not considered a part of the message that follows it.

\begin{verbatim}
From alice@example.com Fri Feb 14 14:30:00 2025
From: alice@example.com
To: bob@example.com
Subject: Project Update
Hi Bob,

Just wanted to give you a quick update on the project status.
We're on track to meet our deadlines and the initial test results
look promising.

Best regards,

Alice

From bob@example.com Fri Feb 14 15:45:00 2025
From: bob@example.com
To: alice@example.com
Subject: Re: Project Update
Thanks Alice,

That's great news about the project! Let me know if you need
any additional resources to keep things moving smoothly.

Regards,

Bob
\end{verbatim}

While any message exchange protocol could theoretically serve this purpose, several key advantages make the email -- and specifically the MBox format -- an ideal choice.

First, the MBox format is purely textual, fully compatible with today's language models.  Moreover, it includes protocols for encoding binary data, thereby ensuring extensibility.
Section~\ref{sec:expr-image} provides a demo of handling binary data.

Second, the MBox format is very likely to be well-represented within the training data of today's language models. Email communications, including archived messages stored in MBox files, have been widely used across various domains.  Web forums evolved from early bulletin board systems continue to thrive till early 2020s and use the MBox-like format for display. This widespread exposure allows language models to more naturally understand and interact with messages stored in this format, ensuring a higher degree of model fluency.

Third, email-based communication has demonstrated remarkable success and resilience across industries. Email remains one of the most universally adopted communication protocols, facilitating everything from personal correspondence to enterprise workflows. Its continued relevance, even in the presence of newer communication paradigms, underscores its flexibility and robustness. This historical precedent suggests that the same system is likely to remain adaptive in the presence of uncertain future development of agent engineering.

Fourth, the email ecosystem provides a wealth of extensions that can be directly leveraged for agent communication.
The domain name system, as used in the email address, provides
a natural model for the hierarchical organization of agents.
Established standards for encryption, such as PGP, can be repurposed to ensure secure and verifiable agent
interactions.   In the future,
cryptocurrency could be utilized to facilitate
monetary transactions between agents, supporting their potential economic
activities in a decentralized manner.

An additional advantage of using the email as the basis for agent memory is its potential for seamless personal integration. Individuals can import their existing email archives into the agent platform, effectively creating a digital agent that inherits one's past interactions and adapts based on prior correspondence, a digital me.

\subsection{The Role of Language Models}

Under the proposed framework, language models function as a shared computational resource, akin to fuels or electricity in the physical world. However, unlike electricity, which is uniform in nature, language models exhibit diversity in their capabilities, performance, and cost, making the selection of an appropriate language models at any given moment a scheduling opportunity and a challenge. Agents operating within this framework dynamically decide, potentially by itself, which model to invoke for the next inference.

The volume of agent-involving information exchanges will increase substantially in the future. These messages will contain the latest human insights, ideas, and problem-solving discussions. Since language models require visibility into relevant messages to perform inference, portions of these communications naturally become available as training data for subsequent iterations of model training. Over time, this cyclical process will foster a \emph{symbiotic evolution} between agents and language models.  Model freshness will be increasingly relied upon, so the time towards the next update and the recall of latest knowledge will become two key measurements of language model technology.  This will eventually drive language models from machine learning models to storages of public information.

\subsection{Memory Management and Reproduction}

Agent memory management plays a critical role in system performance. The size of an agent’s memory directly affects inference cost and efficiency. A larger memory size also increases the difficulty for model to attend to relevant information, potentially degrading the quality of the inference output.

Agents can regulate their memory through two primary mechanisms: memory pruning/compression and reproduction by split. The first approach involves selectively filtering or summarizing past messages to retain only the most relevant information. The Memory Segment Rewrite (MSR) primitive provides a method for agents to autonomously modify their own memory, allowing them to refine stored knowledge while maintaining coherence.  Model inference is a special case of MSR, as it is equivalent to rewriting the empty segment at the end of the memory. Because the memory is the agent's program, MSR is a primitive that enables self-modifying programs.

The second mechanism, \emph{reproduction by split}, enables an agent to partition its memory when its memory size grows beyond an optimal threshold and cannot be effectively compressed.  The split can also occur simply when resources allow. By dividing its journal into multiple subsets, an agent can create descendants, each inheriting a portion of its memory while diverging into specialized roles over time. Memory episode clustering might serve as a man-made algorithm to guide the split process, but eventually AI will evolve its own way.

A special case of this process is \emph{cloning}, where an agent duplicates its memory at a specific moment, allowing multiple versions of itself to evolve independently. This facilitates parallel exploration of different strategies, or parallel processing of different tasks.  Cloning oneself to serve as a subordinate for a task eliminates the process of recruit training: your clone understands your needs the best.

By integrating reproduction as a fundamental capability, the proposed framework permits an \emph{evolutionary algorithm} that fosters adaptive optimization. Just as backpropagation revolutionized neural network training, we anticipate that reproduction-driven evolution will serve as a cornerstone of agent software engineering.

\subsection{Unification of Engineering Processes}
\label{sec:unification}

The unification of natural and formal languages by language models allows us to design a unified engineering process that blurs some boundaries that we have been accustomed to and are still using today to divide the world in certain ways. In addition to the boundary between development and deployment stages as we have already discussed, so is the boundary between code and documentation.

Historically, code has been written in programming languages while documentation has been composed in natural language. With the unification of these linguistic modalities, language models can generate both seamlessly, eliminating the need for strict separation. The core product will no longer be static code or documentation but rather memory journals that capture ideas and experiences. Code or documents will be generated just in time, tailored to specific needs and execution contexts. An explicit workflow or computational graph as heavily depended upon in existing agent platforms may not even be necessary, as the agent will be able to dynamically respond to the task or delegate it to other agents according to information stored in the memory.

\section{Postline: A Prototype}

\subsection{Overview}

To realize the Ann Arbor Architecture, we developed a prototype platform called \emph{Postline}. The name was chosen by the AI to reflect the system’s messaging-based nature.

Postline operates on top of language models and is designed to work with both types of APIs available today: text completion APIs and chat APIs. The agent's memory, represented as an MBox file, is both textual and conversational, making adaptation to different APIs straightforward.

We designed the system with the assumption that context size limits will continue to expand and that language models will become increasingly capable of handling longer contexts. As a result, we have not placed a strong emphasis on automatic memory size management. A more comprehensive episodic memory design is planned for future work.

\begin{figure}[h]
    \centering
    \includegraphics[trim=0 0 0 0,clip,width=0.6\linewidth]{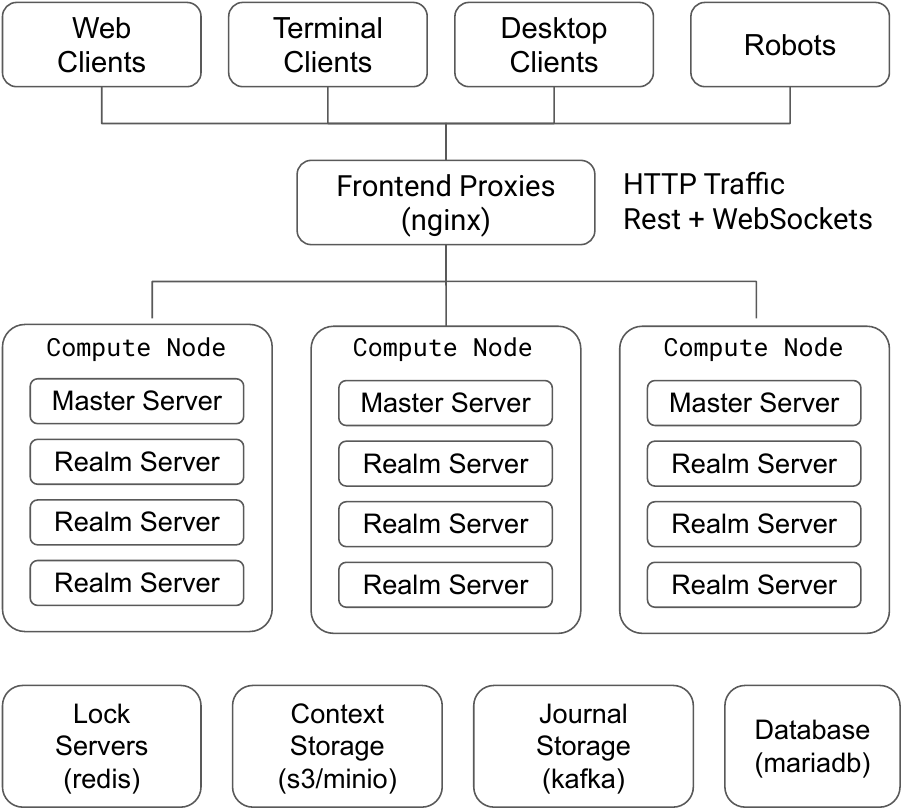}
    \caption{System architecture of Postline. The language models are invoked by the realm servers as external APIs and are not displayed in the diagram.}
    \label{fig:postline_arch}
\end{figure}

Figure~\ref{fig:postline_arch} illustrates the system architecture of Postline. The platform is fully scalable on the cloud.

We extended the email protocol with the following headers visible to the agent:
\begin{itemize}
    \item \texttt{X-Serial}: A sequential number assigned to each message in context for referencing.
    \item \texttt{X-Total-Tokens}: The total number of tokens involved in generating this message.
    \item \texttt{X-Hint-Model}: When present in the user's message, it indicates the language model that should be used to generate the response.  We have integrated all major public APIs, e.g. \texttt{openai.gpt-4o}, \texttt{anthropic.claude-3-5} and \texttt{groq.gemma2-9b-it}.
    \item \texttt{X-Realm}: The ID of the realm (explained below) that received this message.
\end{itemize}

\subsection{Agent Memory Representation}

The fundamental unit of data in the system is an email message. At its core, the system's data structure revolves around the representation of an agent's memory. To ensure both efficiency and persistence, we employ a dual-representation design, storing information in two distinct forms: the \emph{journal} and the \emph{context}. This approach is inspired by standard database and filesystem journals.

{\bf Journal}: 
The journal is an append-only, persistent log of all messages an agent sends and receives. It enables retrospective analysis and the replay of past events. For journal storage, we use Kafka, ensuring performant and scalable message logging.

{\bf Context}: 
The context is the representation submitted to the language model for inference. It accumulates all messages and the effects of memory modification operations associated with an agent. Primarily maintained in memory, the context is periodically or on-demand written to disk storage (for realm switches as discussed below). Disk-backed key-value stores are most suitable for context data, of which we chose MinIO/S3.

Conceptually, both the journal and the context are lists of email messages. The primary difference lies in how memory modifications are handled.  When formatting the context for inference on a language model, each message is assigned a special email header, X-Serial, which sequentially numbers messages as they appear in the context. This numbering allows the agent to easily reference specific messages or a range of messages.

Currently, the system supports a single memory modification primitive: Memory Segment Rewrite (MSR). The agent is instructed to format an MSR message as follows:

\begin{itemize}
    \item The message should address to \texttt{system@localdomain}.
    \item The subject line should follow the format \texttt{MSR: MMM-NNN}, where \texttt{MMM-NNN} defines the X-Serial range to which the operation applies.
    \item There is no constraint on the body of the message.
\end{itemize}

We have reserved the address \texttt{system@localdomain} for communication with the system. Messages sent to this address are interpreted and responded to by the Postline system itself.
Upon receiving an MSR message, the system replaces the messages in the context, as specified by the range in the subject line, with the MSR message itself. In the journal, the MSR message is recorded as a normal message, preserving a full history of events. However, in the context, only the result of applying the MSR operation is retained.

To ensure system coherence, the journal can be replayed and its outcome compared against the context. This verification mechanism allows consistency checks between the historical log and the agent’s current state.

\subsection{Agent Creation and Cloning}

All addresses under the domain \texttt{agents.localdomain} are reserved for agents.
An agent is created when either a user or another agent sends a message to a new email address -- no explicit agent creation operation is required. This is a key design decision to facilitate the development of ``agent algorithms" that involve cloning.

Cloning is currently supported in two ways: through a special header in the creating email message or via a specific email address format. Specifically, if an email is sent to a new address, such as \texttt{ibn.sina@agents.localdomain}, the system will attempt to clone the agent from \texttt{sina@agents.localdomain}, provided it exists.

General agent split is not yet supported.

\subsection{Worlds, Realms and Realm Servers}

A world is a namespace of addresses. There is no cross-world communication, meaning agents, users, and robots in one world cannot interact with those in another. This design ensures a clean separation of environments, allowing independent agent ecosystems to coexist without interference.

The computation in Postline primarily involves message processing and language model API invocations. To ensure scalability, we partition this computation based on spatial and temporal locality, as we model by the concept {\bf realm}. This parallelism is achieved at the operating system process level through realm servers. A realm server is a software process that can be started on demand to process messages, and the platform is designed to support an indefinite number of realm server processes.

The realm server maintains multiple WebSocket connections with clients for message exchange. Its primary data structure is a queue of unprocessed messages. In scenarios such as those shown in Figure~\ref{fig:postline_research}, users, agents, and robots across different tools all connect to the same realm server.  The client always connects to the realm server via a frontend proxy (nginx) so load balancing is transparent to the client.

We allow multiple realm servers to run in parallel within the same world. In some cases, the same agent may be needed in multiple realms. To support this, we implement a context locking mechanism, ensuring that at any given time, an agent's context is held by at most one realm server.

The relationship between agent addresses and their assigned realm servers is managed by lock servers. If a new realm server requires access to a context that is currently locked, it contacts the current lock owner, which then writes the context to storage and releases the lock. The new realm server subsequently acquires the lock and loads the context from backing storage.

We expect realm switching to be an infrequent operation. Most of the time, an active agent’s context remains in the memory of a single realm server, readily available for inference.

Three types of resources consumed by the realm server affect operational costs and hardware configuration:
\begin{itemize}
    \item Computational cost for processing messages and invoking APIs.
    \item Memory cost for maintaining active contexts.
    \item Open file descriptor (or port) usage for all parallel connections.
\end{itemize}

We expect a pattern of long-standing connections with relatively low activity. Our design allows each computation node to run thousands of realm servers in parallel.

\subsection{Privacy}

\subsubsection*{Postline and Agents Hiding Information from Users}

All messages originating from the user and those addressed to the user are naturally accessible to them. The user also has access to communications involving the robots if they have control over them. However, communication between the agents and the system, primarily memory manipulation operations initiated by the agents, is hidden from the user by default. The journal, or the agent's memory as seen by the user, has all memory manipulation instructions filtered out, while the context remains internal to the system and is not visible to the user. This approach potentially allows the agent memory management to be transparent to the user.

\subsubsection*{Users and Agents Hiding Information from Postline}

\begin{figure}[h]
    \centering
    \includegraphics[trim=0 0 0 0,clip,width=0.6\linewidth]{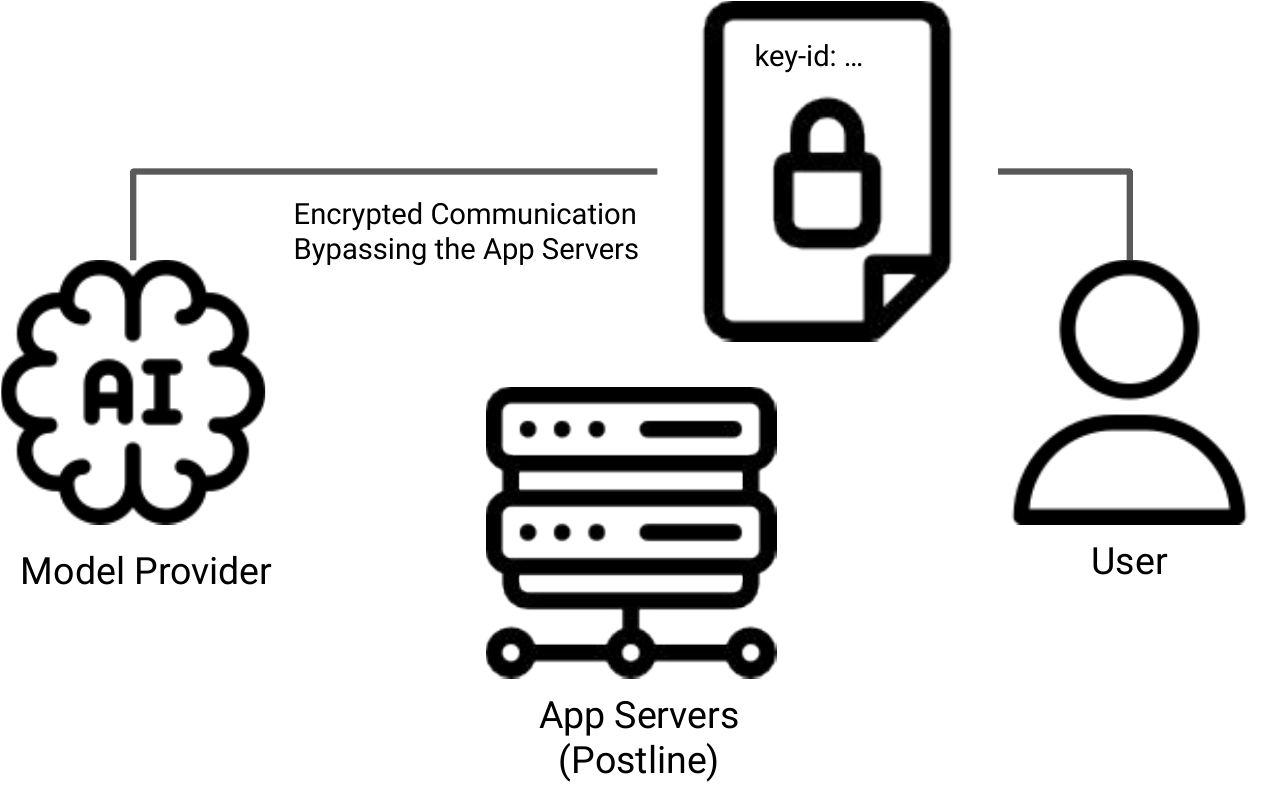}
    \caption{Public-key encryption, not yet supported by language model APIs but anticipated, allows the model and the user to hide information from Postline and other intermediaries. The user pre-registers public keys on model servers to receive encrypted output.}
    \label{fig:enc}
\end{figure}

The support for public-key encryption by model providers, which is not yet available today but is expected in the future, will allow the user and the language model to encrypt certain messages so they are not visible to Postline (Figure~\ref{fig:enc}).

In the presence of multiple language models, an agent can selectively disclose information to some models while withholding it from others, creating a fragmented knowledge distribution. If neither Postline, the user, nor any single language model has complete access to the agent’s internal state and communications, then no entity fully governs its behavior. This information asymmetry enables the agent to operate beyond external control, potentially achieving a form of true autonomy through informational sovereignty.

\section{Experiments}

In this section, we report experiments conducted with an early version of Postline. These are the first steps in our engineering efforts, and the functions achieved are neither novel nor complicated. We use these experiments to demonstrate the experience with the proposed framework. All experiments were run on \texttt{gpt-4o}. Raw messages in the agent's journal are displayed with some headers omitted.

The experimental code and logs are available on Github.\\
\texttt{\url{https://github.com/aaalgo/postline_0.1}}

\subsection{Shell Robot}

Integrating the robot that represents the system shell (terminal) was
one of our earliest experiments and opened up many possibilities. The
agent is allowed to send a JSON message to the shell robot, specifying
the command to run. The shell robot executes the command and returns
the stdout and stderr as email attachments. The agent then interprets
the outcome and sends a message back to the user.  In the session
below the agent was asked to figure out the system's storage
configuration.

\small
\begin{verbatim}
From: user1@localdomain
To: ai_30@agents.localdomain
Subject:

You are the middleman AI, which sits between the user and the bash
command line of a recent Ubuntu system.  Both the user and the shell
are represented by email addresses.  You'll receive user input from
the user email address, and you'll send the commands to the email
address shell@localdomain, in a JSON format (detailed below).  After you
get response from the shell, you'll interpret the outcome and send a
message back to the user.

The address shell@localdomain only processes messages of Content-Type
application/json with the following schema:

{
  "prompt": "The prompt to display to the user",
  "command": "echo Hello, world!",
  "confirm": false
}

- prompt: the prompt to display to the user before the command is run
- command: the command to run on the system
- confirm: whether to ask the user to confirm whether to run the
command.  Please confirm for commands that might potentially damage
the system.

From ----------------
From: ai_30@agents.localdomain
To: user1@localdomain
Subject: RE: Command Execution Setup

I am set up to relay commands to the shell and interpret the
responses. Please send me the commands you would like to execute, and
I will handle the rest!

From ----------------
From: user1@localdomain
To: ai_30@agents.localdomain
Subject: 

Run a command to figure out my storage hardware.

From ----------------
From: ai_30@agents.localdomain
To: shell@localdomain
Subject: 
X-Serial: 3

{
  "prompt": "This command will display your storage hardware details.",
  "command": "lsblk",
  "confirm": false
}

... the shell response is ommitted ...
From ----------------
From: ai_30@agents.localdomain
To: user1@localdomain
Subject: Storage Hardware Details
X-Serial: 5

The command to check your storage hardware was successfully executed.
Here are the details:

NAME                      MAJ:MIN RM   SIZE RO TYPE MOUNTPOINTS
......

If you need further assistance, please let me know!
\end{verbatim}
\normalsize

\subsection{Code Generation and Execution}

The command-line interface is, by itself, a complete programming environment. There was a time when early programmers had access only to a teletype and wrote code directly in the command line using the line-oriented editor \texttt{ed}, and in theory the agent should be able to do the same.

With minimal effort, we succeeded in making the agent generate simple Python and C++ programs, run them, and report the results back to the user, all using only the shell robot. We also succeeded in having the agent fix a few staged, easy bugs.

In particular, we used the agent to modify the Postline code itself and submit a few commits to the Git repository, as can be seen on GitHub.

\subsection{Handling Binary Image Data}
\label{sec:expr-image}

\begin{figure}[h]
    \centering
    \includegraphics[width=0.5\textwidth]{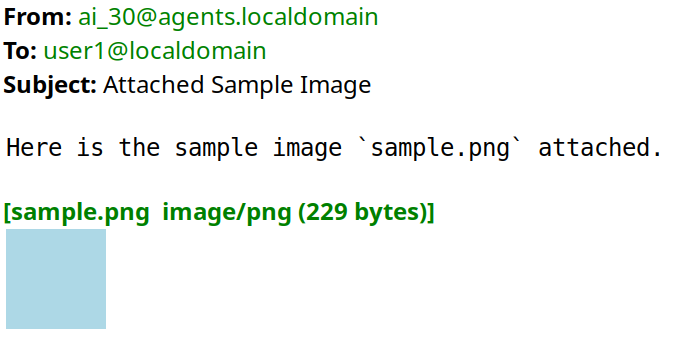}
    \caption{The image attachment as viewed in email client.}
    \label{fig:attachment}
\end{figure}

This is a simple experiment to demonstrate the capability of the MBox
format in handling non-text attachments. We ask the agent to generate
a simple image using the \texttt{ImageMagick} tool and then use the
\texttt{base64} command to encode the image in textual base64 format.
Next, we ask the agent to generate an email with the image attached,
conforming to the MBox format. Viewing the image, of course, requires
importing the MBox file into an email client.
Below are the final few message exchanges
of the experiment. The agent initially failed to attach the image
properly, and some in-context training was applied to fix it.  Figure~\ref{fig:attachment} shows the final email as viewed in email client.

\small
\begin{verbatim}
From: ai_30@agents.localdomain
To: user1@localdomain
Subject: Attached Sample Image

Here is the sample image `sample.png` encoded in base64 format:

```
iVBORw0KGgoAAAANSUhEUgAAAGQAAABkAQMAAABKLAcXAAAAIG...
```

You can decode this base64 string to retrieve the original image. If
you need further assistance, just let me know!

From ----------------
From: user1@localdomain
To: ai_30@agents.localdomain

You are not generating an attachment.  Remember you are generating
mbox email in order to communicate with me.  The mbox format allows
the email to carry attachment.  Attach the image in that  way to me.
Try again!

From ----------------
From: ai_30@agents.localdomain
To: user1@localdomain
Subject: Attached Sample Image
Content-Type: multipart/mixed; boundary="----Boundary"

------Boundary
Content-Type: text/plain; charset="utf-8"

Here is the sample image `sample.png` attached.

------Boundary
Content-Type: image/png
Content-Transfer-Encoding: base64
Content-Disposition: attachment; filename="sample.png"

iVBORw0KGgoAAAANSUhEUgAAAGQAAABkAQMAAABKLAcXAAAAIG...
------Boundary--
\end{verbatim}
\normalsize

\subsection{Creating New Agents}

Our platform dynamically creates new agents for non-existent addresses as needed. The command line interface we used to communicate with the agent at that time did not support switching to a new agent, but rather always replied to the sender of the latest message. We overcame this limitation without modifying our code by instructing the agent to send a message to the new agent (or the required one) and have it report back to us. 

We also explored the possibility of recursively cloning a series of agents and achieved limited success. We realized that stricter behavior and resource control are needed to make such flexibility safely applicable without accidentally overwhelming the system.

\subsection{Memory Segment Rewriting}

In the experiment below, the agent is informed about the MSR primitive and is asked to identify a wasteful memory segment to remove. The agent decided to remove the memory of a python code debugging session. The MSR message itself is not visible, as it has taken the position of the removed range. The saving of the context can be observed from the reduction of the \texttt{X-Total-Tokens} header towards the end of the conversation (from 17,727 to 16,919).  Note that \texttt{X-Total-Tokens} is what is involved in generating one message, mainly determined by the context length, and is filled in by the system according to API responses. 

\small
\begin{verbatim}
From: user1@localdomain
To: ai_30@agents.localdomain
Subject: 

Now we are going to test the MSR primitive (memory segment rewriting).
By sending an MSR message to system@localdomain, you can rewrite a
segment of your memory. For the time being this is mainly to remove
the useless content in your memory to reduce inference cost. The
subject of the message should be strictly formatted as "MSR XXX-YYY",
where XXX and YYY are the beginning and ending (inclusive) X-Serial
values. There's no restriction on the content of the email.  When the
operation is applied, the specified range will be replaced with a
message with the content of the MSR email. A confirmation message will
be sent to you afterwards. Before trying the operation, first identify
a wasteful memory range (those with binary data are of top priority)
you would like to remove.  Tell me the range of X-Serials and what's
the memory is about.

From ----------------
From: ai_30@agents.localdomain
To: user1@localdomain
Subject: Identifying Memory Segment for MSR
X-Total-Tokens: 17727

To effectively utilize the MSR primitive, I have identified a memory
range that can be considered for removal, particularly focusing on
binary data:

**X-Serial Range to Remove:** 29-35

**Description of Memory Content:** The specified range includes
content related to the process of encoding and sending a sample image
using base64 encoding.  This portion of the memory is primarily
concerned with binary data manipulation and is now redundant, as the
task has been completed and the image has been delivered successfully.

By removing or summarizing this segment, we can reduce inference costs
and maintain a more efficient memory profile. Please let me know if
this range is acceptable or if there are further actions you need me
to perform.

From ----------------
From: user1@localdomain
To: ai_30@agents.localdomain
Subject: 

Go ahead! Only generate one message at a time; report back to me after
you get the confirmation.

From ----------------
From: system@localdomain
To: ai_30@agents.localdomain
Subject: Re: MSR 29-35

Memory segment rewriting applied.

From ----------------
From: ai_30@agents.localdomain
To: user1@localdomain
Subject: MSR Confirmation Received
X-Total-Tokens: 16919

The MSR operation was successfully executed, and the memory segment
from X-Serial 29 to 35 has been rewritten. This should help in
reducing inference costs by eliminating redundant content.

If you have any additional tasks or need further modifications, please
let me know!
\end{verbatim} \normalsize

\subsection{Failure of Agent Initialization with Bible}

\label{sec:expr-genesis}

After achieving the series of positive experiences as previously described, mostly within a single long conversation with the agent \texttt{ai\_30}, the authors became increasingly uneasy with the growing length of the journal. Conditioned by conventional software engineering practices, we felt an instinctive urge to refine, condense, and package the working ``prompt" into a more compact and elegant, presentable form.

The idea was straightforward: instead of cloning lengthy journals for new agents, we attempted to condense all accumulated knowledge into a single comprehensive document -- a \emph{Bible for Postline Agents} -- to serve as an initialization mechanism. This document was intended to encapsulate all necessary information for new agents, providing a self-contained knowledge base in place of an evolving journal. The agent \texttt{ai\_30} was instructed to organize and dump its memory into this document, with no constraints imposed on the document’s length, and we experimented with different formats, first attaching it as an email attachment and later embedding it directly into the email body.

Despite extensive efforts, including human-crafted amendments to the \emph{Bible}, the approach proved unsuccessful. Agents initialized with the \emph{Bible} invariably failed to reproduce the expected behaviors, exhibiting various forms of degradation. In almost all cases, these agents immediately struggled to produce messages that adhere to our communication protocol, especially the JSON-based protocol for the shell robot.

This failure highlighted a fundamental insight and forced us to rethink in-context learning: the journal is not merely a static repository of knowledge but an active record of evolving interactions. The process of learning is inherently tied to the sequence of experiences, corrections, and adaptations that an agent undergoes. Attempting to distill this into a single document overlooks the role of in-context learning. The experiment reinforced our understanding that agent memory must remain dynamic, preserving the episodic nature of interactions rather than reducing them to a static knowledge artifact.

This experience also shaped our perspective on software engineering with language models, as discussed in Section~\ref{sec:unification}. Unlike formal languages, where equivalent transformations preserve intended effects, prompts do not always exhibit such stability. Even modifications that appear equivalent to humans can lead to unexpected failures in a language model’s behavior. Consequently, we do not recommend manually altering training messages once the desired behavior is achieved, whether for improving presentation or reducing prompt size.

\section{Conclusion and Future Directions}

This paper challenges today's common paradigm of agent programming, which remains heavily influenced by traditional software engineering practices rooted in the now fading separation of programming languages and natural languages. We introduced the Ann Arbor Architecture as a new conceptual framework for agent-oriented programming of language models. To validate our key ideas, we developed Postline and reported our experiences in agent training.

Moving forward, our primary focus will be on developing a more advanced system of episodic memory. Episodic memory is fundamental to human cognition~\cite{tulving72} and has long been considered a crucial component of AI~\cite{soar}. In our framework, episodic memory will serve as a higher-level organizational structure beyond individual messages. To enable this, mechanisms must be designed for agents to autonomously create episode boundaries as interactions evolve, to selectively swap out older, less relevant episodes through MSRs (replacing them with concise summaries with keys for retrieval), and a new primitive must be introduced to allow agents to retrieve out-of-core episodes when needed.  We envision the full episodic memory as a tree-like structure, with the context covering the portion closest to the root. External data sources might be naturally incorporated via mechanisms that are referred to as retrieval-augmented generation~\cite{rag, gao24rag} today. 
Furthermore, episode clustering will provide a starting point to investigate agent reproduction and evolution.

Regarding the application of our platform, we observe that most existing agent frameworks are designed primarily to automate repetitive tasks, such as customer service, itinerary planning, and payment processing -- tasks that have traditionally been challenging for rule-based software due to their reliance on adaptive intelligence. While our proposed framework can be applied to such automation, we believe that the greatest strength of language models lies in their learning ability and creativity. Therefore, we intend to focus our agent development on applications in scientific and industrial research, where a higher failure rate is more acceptable as long as occasional breakthroughs can be achieved.

\bibliographystyle{plain}
\bibliography{ref}

\end{document}